# A BIM-enabled, Agent-based Discrete-event Simulation Platform for Robotic Studies: A Method based on Graph Theory

Ping Xu[1], Xinghua Gao[1*]
[1]Myers-Lawson School of Construction, Virginia Tech, Blacksburg, 24060, United States.

[*]**Corresponding Author**, Myers-Lawson School of Construction, Virginia Tech, 320J Hitt Hall, Blacksburg, 24060, United States; Email: xinghua@vt.edu

**Abstract:**
Indoor robots are increasingly employed for facility management tasks such as cleaning and inspection. These applications primarily rely on navigation and can be effectively supported by pre-defined routes or perception-driven Simultaneous Localization and Mapping (SLAM) techniques. However, more complex tasks, such as locating and repairing leaking pipes, require not only navigation but also access to building information, including the location, geometry, material, and operational attributes of components. Existing navigation approaches provide only limited environmental understanding and cannot readily supply such information. In contrast, Building Information Modeling (BIM) contains rich geometric, semantic, and operational information that remains largely underutilized in robotic applications. This study proposes a BIM-enabled, agent-based simulation platform for knowledge-driven indoor robot navigation and operation planning. Within the framework, indoor environments are discretized into grid cells that are mapped to graph nodes and classified as target, obstacle, or regular nodes according to their spatial relationships with building elements. Traversal costs are assigned to edges connecting neighboring nodes, enabling graph-theoretic algorithms to compute efficient and collision-free navigation paths while avoiding obstacles. Simulation results demonstrate that the proposed graph representation enables efficient and collision-free navigation. A key limitation associated with coarse discretization—overlap between target-occupied and obstacle-occupied cells—is identified and mitigated through grid refinement, improving spatial accuracy and path feasibility. The proposed platform supports virtual evaluation of robotic operations prior to deployment and provides a foundation for BIM-informed robotic systems in facility management.

## 1. Introduction

Robotic systems have increasingly been deployed to perform simple and repetitive tasks in indoor environments, such as inspection and cleaning. However, the intelligence of most existing indoor robots remains limited. Current indoor navigation methods predominantly rely on pre-defined paths or Simultaneous Localization and Mapping (SLAM) techniques to plan navigation routes. While these approaches enable robots to move autonomously, they require continuous perception, localization, and real-time computation, resulting in substantial computational burdens. More importantly, robots operating under these paradigms lack a comprehensive understanding of the indoor environment and access to task-related operational information, such as the specifications and manufacturer information of a leaking pipe. This limitation fundamentally constrains their decision-making capabilities and operational intelligence.

In typical indoor navigation scenarios, robots perceive the environment locally and incrementally construct maps through onboard sensors. This sensor-driven, perception-centric paradigm forces robots to repeatedly infer spatial relationships that are already known during the building design and operation phases. However, for buildings that are designed, constructed, and managed using Building Information Modeling (BIM), robots can obtain maps based on rich, structured, and semantic building knowledge from existing BIM models, eliminating the need to reconstruct indoor environments. BIM models provide detailed and semantically rich representations of indoor environments, including the geometry and spatial relationships of building elements, as well as the locations of furniture and equipment such as sofas, tables, and other obstacles. Beyond geometric information, BIM models also store operational and maintenance (O&M) data that are highly relevant to indoor robotic tasks. Leveraging this existing information offers a promising opportunity to shift robot navigation from a purely perception-driven approach to a knowledge-driven paradigm.

In this study, we propose a BIM-enabled, agent-based simulation platform in which a robot is modeled as an intelligent agent operating within a BIM environment. A graph-based representation of the indoor space is constructed by discretizing the environment into grid cells, each of which is mapped to a graph node. Based on their spatial relationships with building elements and task-related objects, these nodes are further classified as target nodes, obstacle nodes, or regular nodes. Edges between neighboring nodes are assigned traversal costs according to spatial relationships and navigational constraints, while obstacle nodes are treated as non-traversable. By applying graph theory, the robot is able to compute collision-free navigation paths that reduce travel cost while avoiding obstacles. Within the proposed simulation platform, the robot can access comprehensive indoor knowledge directly from the BIM model, including geometric, spatial, and O&M-related information. This enables the robot to plan navigation paths efficiently and purposefully, reaching target locations efficiently while reducing unnecessary computation. The platform allows robotic operations to be simulated and evaluated in a virtual BIM environment before deployment, thereby enhancing navigation intelligence and operational reliability.

In addition, the graph-based formulation enables a clear separation between spatial representation and decision-making algorithms. By abstracting the indoor environment into a structured graph, navigation problems can be solved independently of the underlying geometric complexity. This abstraction not only simplifies graph-based computation but also allows the integration of different optimization strategies without modifying the environment representation. As a result, the proposed approach provides a scalable foundation for extending indoor robotic navigation from simple path planning tasks to more complex decision-making scenarios.

## 2. Literature Review

This research employs a literature review to explore the capabilities of BIM that can be leveraged to construct the simulation engine and to explore how the graph theory can be used for indoor navigation. To achieve this, the literation review is conducted from two perspectives: 1) BIM for developing a simulation engine; and 2) graph theory for indoor navigation.

### 2.1 BIM for developing a simulation engine

BIM has been widely applied as a simulation engine in construction. Ahn et al. [1] used 3D BIM simulation to represent the hazard conditions on sites for developing effective safety training and promoting workers' safety awareness. Torres-Calderon et al. [2] proposed a method that uses text mining and machine learning to achieve 4D BIM simulations automatically for optimizing construction planning and scheduling. Pérez & Bastos Costa [3] evaluated the utility of 4D BIM simulations for reducing transportation waste. With 4D BIM, users can create and view animations that 3D objects move in the 4D simulation of construction activities. Alzarrad et al. [4] presented a guideline to identify errors in construction scheduling for renovation projects by applying 4D BIM simulation, which can reduce project cost and duration.

BIM provides the data required by simulation software and simulation engines to simulate building performance and construction processes. Rüppel and Schatz [5] developed a BIM-based serious game for fire evacuation simulation, in which the BIM model provides the virtual building environment for simulating different fire evacuation scenarios. Kim et al. [6] developed a BIM-based thermal energy simulation engine that uses BIM data, such as the parameters of building components, to obtain component-level energy performance. Kota et al. [7] proposed a method to integrate BIM and daylighting simulation software such as Radiance and DAYSIM to simulate the daylighting performance of buildings. They developed add-in programs to translate BIM models, which contain the simulation required data, into Radiance and DAYSIM input files. Osorio-Sandoval et al. [8] proposed a framework of generating construction simulation models by integrating BIM and game engine. They developed an interface called Constructable Element to translate BIM data into the simulation models for the framework. Hosseini and Maghrebi [9] proposed a framework that integrates the 4D-BIM and Social Force Model simulation engine for simulating the evacuation performance of fire emergency. In this framework, 4D-BIM provides facility's information for simulation.

In this research, BIM serves as the data platform for providing building element information, such as the location of individual building elements and the overall layout of the building.

## 2.2 Graph theory for indoor navigation

Graph theory is a branch of mathematics that focuses on the properties and applications of graphs [10]. A graph is a collection of objects, called nodes (or vertices), connected by edges (or links) [11]. Graphs are widely used to represent relationships, structures, and processes in various fields, including computer science, engineering, biology, social sciences, and transportation systems. When represented as graphs, complex systems can be efficiently analyzed using graph algorithms such as Dijkstra's algorithm to solve shortest path problems [12]. Likewise, BIM-derived graph representations, together with appropriate graph algorithms, can be employed to support indoor robot navigation by generating efficient paths that enable robots to reach target locations while avoiding obstacles.

Gilliéron et al. [13] employed graph theory, specifically the link/node representation of a network, as a fundamental component in developing an indoor navigation system. This representation enables the use of classical path-finding algorithms and provides a structured approach to modeling the indoor environment for localization and route guidance. Similarly, Jensen et al. [14] proposed a base graph model that captures the topology of indoor spaces at different levels. Based on this model, an RFID-specific deployment graph was developed to facilitate the construction and refinement of moving object trajectories using RFID readings. This study leverages graph theory to provide a uniform and flexible framework for managing indoor positioning and tracking data, including the deployment of RFID readers. Liu et al. [15] presented a novel indoor navigation system called PILOT, which exploits the peak intensities of ubiquitous unmodulated luminaries to create a virtual graph representation for navigation. The virtual graph serves as a global reference frame for indoor navigation, allowing users to associate themselves with nearby nodes without the need for a pre-deployed floor map or localization system. Fu and Liu [16] proposed a BIM-based navigation graph network to represent all available paths in a building. The navigation graph network was used to model pedestrian navigation and the interaction between pedestrians and exit signs for evaluating building exit sign systems.

Existing studies have demonstrated that graph theory provides an effective framework for representing indoor environments and supporting navigation tasks. Graph representations enable complex indoor spaces to be transformed into structured topological models, allowing efficient path planning and navigation through well-established graph algorithms. However, existing graph-based approaches primarily focus on representing building topology, pedestrian navigation, or indoor positioning. Limited attention has been paid to how graph representations should be constructed specifically for BIM-enabled indoor robot navigation, particularly with respect to representing navigable space, targets, and obstacles. To address this gap, this research develops a BIM-derived graph representation for indoor robot navigation. The graph construction strategy and its underlying design principles are presented in Section 3.2.

## 3. Methodology

### 3.1 Design science and agent-based modeling

In this research, we adopt a design science approach, specifically agent-based modeling, to construct the simulation engine. Design Science is a methodology that facilitates the development of innovative solutions to address real-world problems [17]. In this study, we develop a simulation engine as a novel tool to simulate robot indoor navigation as a foundational robotic operation in the built environment. Agent-based modeling is a technique for modeling and simulating complex systems composed of discrete agents, making it well suited for representing agent behavior and interactions within an environment [18].

This simulation framework consists of three primary modules: 1) the agent module, 2) the environment module, and 3) the module that facilitates interactions between the agent and the environment. In this research, the robot serves as the agent, the BIM model provides the building environment, and robot indoor navigation—including path planning and obstacle avoidance—is simulated through the interaction module between the robot and the building environment. By modeling indoor navigation as a fundamental robotic operation, the proposed framework establishes a foundation for extending the simulation to more complex robotic tasks in future work.

The proposed simulation framework follows the design science paradigm by integrating agent-based modeling, graph-based navigation, and BIM-enabled environment representation into a simulation platform. The agent module models robot behaviors, the environment module provides spatial and semantic building information, and the interaction module enables navigation and decision making through graph-based path planning. These modules establish the overall methodological framework of this research. The following subsections describe the graph-based navigation methodology and the BIM-enabled implementation of the proposed simulation framework.

### 3.2 Graph-based path planning and obstacle avoidance

To achieve path planning and obstacle avoidance, this research adopts graph theory, which represents a system as a set of nodes (vertices) and edges describing the connectivity between them. The BIM-derived indoor environment is converted into a weighted graph, where nodes correspond to discretized spatial units (i.e., grid cells) and edges represent feasible movements between adjacent cells. This graph formulation transforms indoor navigation into a structured optimization problem that can be efficiently solved using established graph search algorithms.

Indoor navigation is modeled using a graph whose nodes represent discretized spatial units rather than building elements. This distinction is fundamental because robots move through space instead of traversing physical objects. Accordingly, building elements are interpreted according to their roles in navigation rather than being directly represented as graph nodes. Target elements define navigation destinations that the robot must reach, whereas obstacle elements define regions that must be avoided.

To accurately represent navigable space, the floor area is discretized into uniform grid cells, each corresponding to a graph node. This establishes a one-to-one relationship between graph nodes and discretized spatial units. Unlike approaches that represent each building element using a single graph node, the proposed representation preserves the spatial occupancy of building elements. For example, if a sofa were represented by only one graph node, obstacle avoidance would prevent traversal through only that single location, while the remaining occupied area could still be incorrectly treated as navigable. Instead, every grid cell occupied by the sofa is classified as an obstacle node, allowing the entire occupied region to be excluded from navigation.

Target elements are modeled differently. Because a target serves as a navigation destination rather than an area that must be represented during path planning, representing every occupied grid cell as a target node would create multiple candidate destinations and unnecessarily complicate shortest-path computation. Therefore, only the grid cell containing the designated navigation destination of a target element is assigned as the target node, ensuring a unique destination for graph search while maintaining a one-to-one correspondence between navigation tasks and graph nodes.

During graph construction, the grid cell containing the designated navigation destination of each target element is labeled as a target node, grid cells intersecting obstacle elements are labeled as obstacle nodes, and all remaining grid cells are treated as regular nodes. Edges are then created between adjacent nodes to represent feasible local movements. To prevent traversal through obstacles, prohibitively high (effectively infinite) costs are assigned to edges connected to obstacle nodes. The resulting weighted graph provides the basis for collision-free shortest-path planning.

The different representations of target and obstacle elements are intentional because they reflect fundamentally different navigation requirements. Each target element is associated with a unique navigation destination, whereas obstacle elements occupy physical space that must be avoided throughout their entire extents. Consequently, the proposed framework separates path planning from task execution. During path planning, graph algorithms operate on the discretized graph using the target node as the graph-search destination. During task execution, the real-world coordinates associated with the designated navigation destination are retained and used as the robot's final destination. The designated navigation destination is task-dependent. For the cleaning task considered in this study, it corresponds to the center of the target element. This dual representation connects the discrete planning space with the continuous execution space while preserving both computational efficiency and navigation accuracy.

Beyond enabling path planning and obstacle avoidance, the proposed graph representation provides several additional advantages. Compared with continuous geometric representations, graphs transform indoor navigation into a discrete optimization problem that can be efficiently solved using existing graph search algorithms. They also naturally accommodate additional navigation constraints, such as restricted areas, safety zones, or task priorities, by modifying node classifications or edge weights without changing the underlying graph structure.

The proposed graph formulation also scales well to large and complex indoor environments. As building size increases, local connectivity between adjacent nodes limits the search space of path-planning algorithms, allowing navigation problems to remain computationally tractable. Furthermore, the graph abstraction separates geometric modeling from algorithm design, enabling different routing and optimization algorithms to operate on the same graph representation without modifying the underlying spatial model.

Another advantage of the proposed grid-based representation is its adjustable spatial resolution. Grid size can be selected to balance computational efficiency and navigation accuracy according to application requirements. Coarser grids reduce the number of nodes and edges, improving computational efficiency, whereas finer grids provide a more faithful representation of the indoor environment and produce more accurate navigation paths.

### 3.3 BIM-enabled simulation framework and system implementation

Further methodologies and tools employed in this research include BIM, Autodesk Revit, and the C# programming language. BIM is a digital modeling paradigm combined with a set of processes for creating, managing, and analyzing building information throughout the building lifecycle, enabling enhanced visualization, simulation, and performance analysis of built environments [19]. BIM models contains rich, structured, and semantic building information, including the geometric properties and spatial locations of building elements, as well as O&M data, all of which can be accessed programmatically through an Application Programming Interface (API).

Leveraging these capabilities, BIM is adopted in this research as a prior source of spatial knowledge for indoor robot navigation. By directly extracting building geometry and semantic information from the BIM model, the robot can obtain an explicit representation of the indoor environment without the need to construct a map from scratch. Compared to conventional perception-based mapping approaches such as SLAM, this BIM-enabled strategy reduces computational burden and allows the robot to acquire spatial cognition of the environment at an early stage. Once the BIM model is integrated with the robotic system, navigation and path planning can be performed based on accurate, building-level information that is consistent with the as-designed or as-maintained environment.

Revit, developed by Autodesk, is software designed for creating BIM models. We use Revit to craft the building model, which serves as the simulation environment. Revit is a potent tool for creating, managing, and sharing digital building models. Furthermore, it offers the Revit API, which enables users to extract location data of building elements such as location coordinates. Such location data is crucial for simulating the placement of agents within the building. During the simulation process, obtaining location data of the building elements interacting with agents is essential, and the Revit API permits the Revit plugin to execute external commands, like simulating interactions between agents and the building environment. C#, a programming language based on the .NET framework, is used to develop the Revit plugin. Since the Revit API is built upon the .NET framework, compatible programming languages like C# can be utilized to develop plugins. These plugins can extract data from building models or execute external commands on building models, such as moving building elements through the Revit API [20]. Fig. 1 illustrates the methods and tools used to achieve the research objectives.

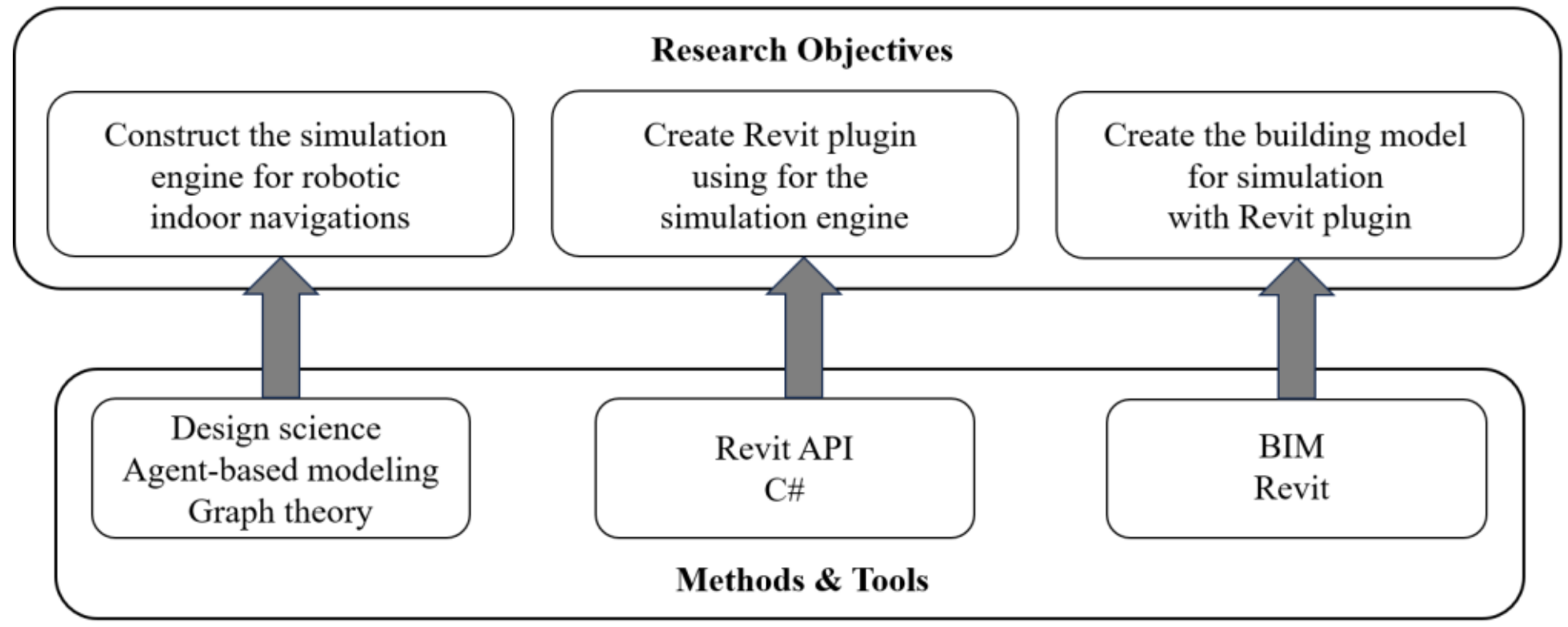

**Fig. 1.** Research methodology.

# 4. System Design and Prototyping

We propose a BIM-enabled simulation platform, as illustrated in Fig. 2, to model robot–environment interactions and support graph-based indoor navigation. The system consists of three main components: the building model, the simulation engine, and the Revit API, which serves as the communication interface between them. The building model provides both geometric and semantic information of indoor environments, while the simulation engine processes this information to generate navigation decisions for robotic agents.

The simulation engine is composed of three modules: the agent simulator module, the building environment module, and the interaction module. A custom Revit plugin was developed to integrate these modules, enabling the simulation of robot operations within BIM environments and facilitating interaction between agents and building elements. Through this integration, the simulation platform is able to directly utilize BIM-based spatial and semantic data without requiring external data conversion.

A key feature of the proposed system is the transformation of BIM-based spatial information into a graph representation, which supports efficient path planning through discretization and abstraction of the indoor environment. This transformation bridges the gap between rich BIM data and computationally efficient graph-based algorithms, enabling scalable simulation of robot navigation tasks.

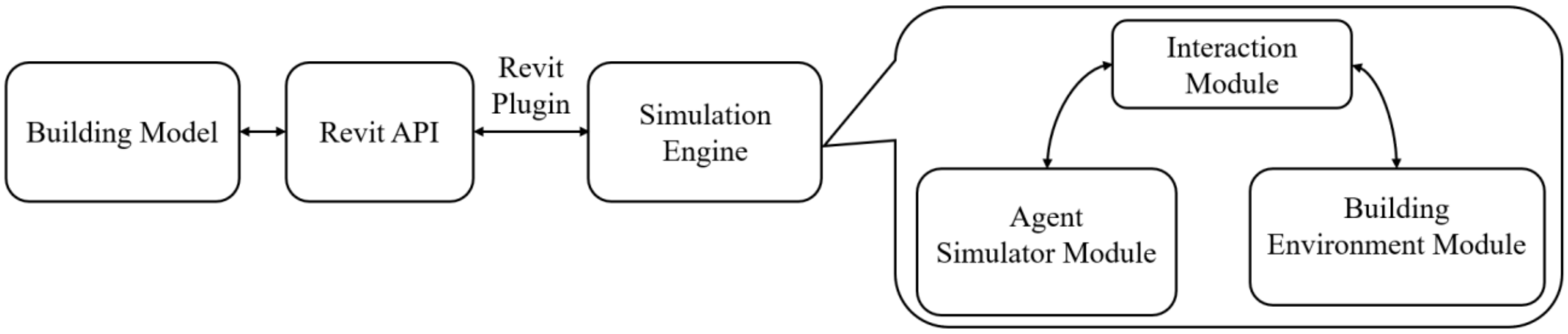

**Fig. 2.** The conceptual framework of the simulation platform.

## 4.1 Modular design for system prototyping

The agent simulator module defines the properties and behaviors of the robot. The robot is modeled with both static attributes (e.g., name, power status, and availability) and dynamic behaviors (e.g., movement speed and navigation actions). To ensure computational efficiency, the robot is abstracted as a point-based agent operating within a discretized environment. This abstraction reduces the complexity of collision detection and motion planning while preserving the essential characteristics required for navigation.

The building environment module represents the indoor space and its associated elements. An IndoorEnvironment class is used to describe the state of the built environment, including spatial regions (e.g., living room), obstacles (e.g., furniture), and targets (e.g., trash). These elements are retrieved from the BIM model through the Revit API, allowing the simulation to leverage both geometric properties (e.g., size and location) and semantic attributes (e.g., element category and family type). This integration ensures that the simulation remains consistent with the underlying BIM model.

The interaction module governs the relationship between the agent and the environment. In this module, graph theory is employed to model navigation, where the robot moves through a discretized spatial representation. The objective is to compute an optimal path that allows the robot to visit all targets while avoiding obstacles. This module integrates the outputs of the agent and environment modules and serves as the core of the simulation engine. In addition, this module supports flexible adaptation to different facility management tasks by modifying the definitions of targets and obstacles based on task requirements.

### 4.2 Graph construction for indoor navigation

The indoor environment is represented as a graph, where nodes correspond to spatial locations and edges represent feasible movements between adjacent locations. To construct this graph, the continuous indoor space is discretized into uniform grid cells. Each grid cell is mapped to a node in the graph, forming a structured representation of the environment. This discretization enables the transformation of complex spatial layouts into a computationally manageable graph structure.

A continuous point (x, y) is mapped to a discrete grid cell index (i, j) using:

$$(i,j) = \left( \left\lfloor \frac{x - x_{min}}{\Delta} \right\rfloor, \left\lfloor \frac{y - y_{min}}{\Delta} \right\rfloor \right)$$

where $\Delta$ is the grid cell size, and $x_{min}$, $y_{min}$ denote the lower bounds of the environment in the global coordinate system.

This formulation enables a consistent mapping between continuous BIM-based coordinates and discrete graph nodes, and supports flexible adjustment of grid resolution. As a result, the same formulation can be applied under different discretization scales (e.g., 1 ft × 1 ft or 0.5 ft × 0.5 ft), depending on the desired level of spatial accuracy.

For the cleaning scenario considered in this study, graph nodes are classified according to the roles of building elements in the task into three types: target nodes, obstacle nodes, and regular nodes. As discussed in Section 3.2, each target element is associated with a designated navigation destination. For the cleaning task, this navigation destination is defined as the center of the target element. Target nodes correspond to grid cells containing the designated navigation destinations of target elements (i.e., the centers of banana peels or water spills in this study), whereas obstacle nodes correspond to grid cells intersecting the projected footprints of obstacles (e.g., furniture). All remaining cells are classified as regular nodes. This classification enables the graph to capture both navigable space and spatial constraints imposed by obstacles.

Edges are constructed between neighboring nodes (e.g., 4-connectivity), and traversal costs are assigned based on node types. Obstacle nodes are treated as non-traversable, while regular and target nodes are assigned distance-based costs. This graph structure supports path-planning algorithms that compute optimal navigation paths for the robot. The use of grid-based adjacency ensures that the resulting paths are computationally efficient to compute and easy to interpret.

To automate graph construction, building elements are retrieved from the BIM model using filtering functions provided by the Revit API. Target elements (e.g., those labeled as “Trash”) and obstacle elements (e.g., furniture) are identified based on their semantic attributes. Obstacle nodes are generated from grid cells intersecting the projected footprints of obstacle elements, whereas target nodes are generated from the grid cell containing the designated navigation destination of each target element. This process enables automatic graph construction directly from BIM data without manual intervention.

Under coarse grid resolutions, a grid cell may simultaneously intersect with both target and obstacle regions, leading to ambiguity in node classification. This issue arises due to the limited spatial resolution of discretization and reflects an inherent trade-off between computational efficiency and geometric accuracy. This issue is addressed in the subsequent section through refinement of the grid resolution.

Beyond addressing this limitation, the proposed graph representation serves as a flexible foundation for integrating different navigation algorithms. Because the graph abstraction is independent of the routing strategy, different algorithms can be incorporated according to task requirements. For example, Dijkstra's algorithm or A* can be applied to compute collision-free shortest paths, whereas optimization- or heuristic-based routing methods can be integrated for multi-target navigation. This flexibility highlights the modularity and extensibility of the proposed framework.

## 5. Experiment

In this section, we developed a building model (as shown in Fig. 3) consisting of a living room, a bedroom, and a kitchen. Two banana peels and two water spills were placed in the living room. The experiment was conducted within this space, where a cleaning robot was tasked with removing the banana peels and water spills while avoiding obstacles, including tables, chairs, and sofas.

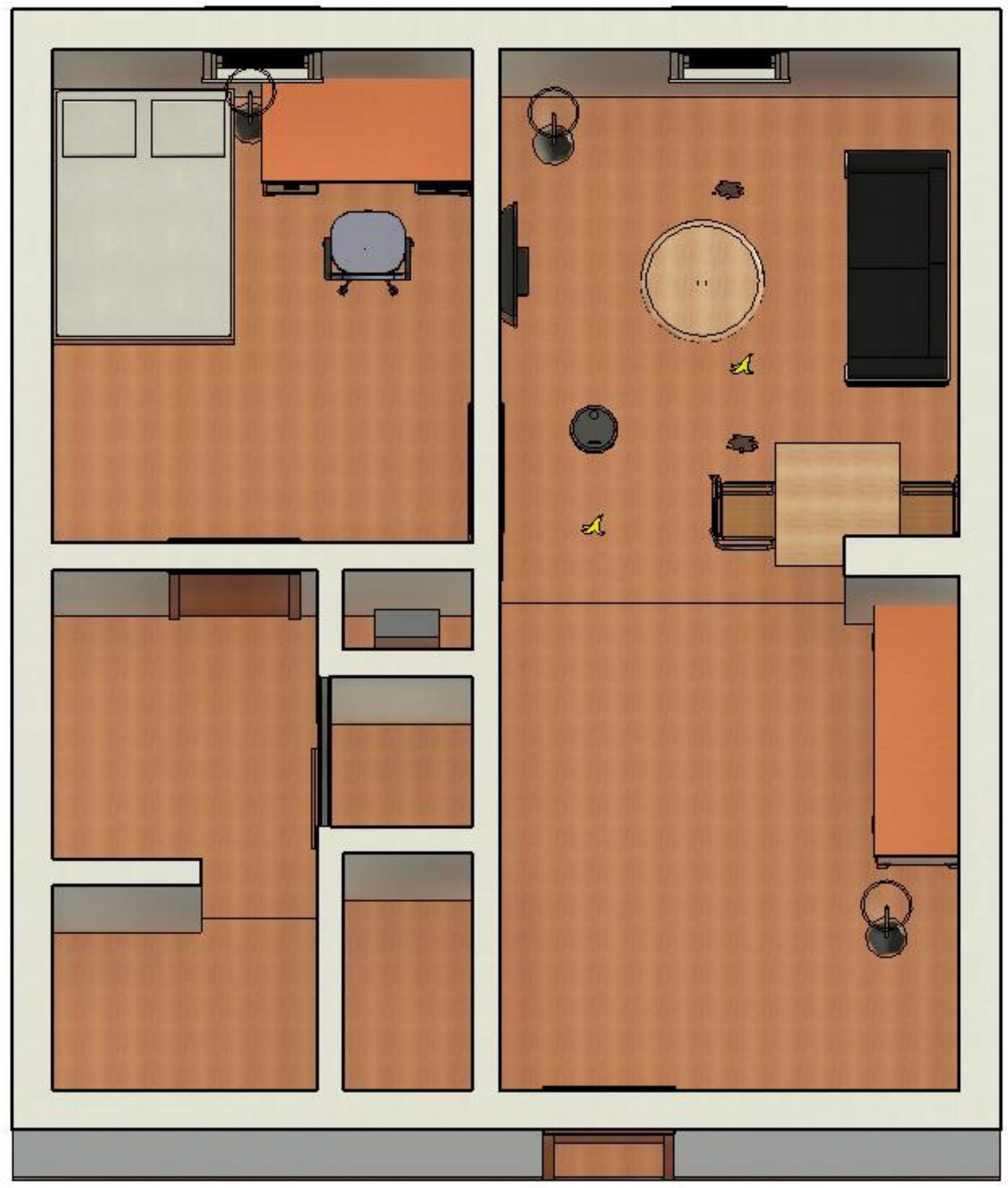

**Fig. 3.** The building model for experiment.

### 5.1 Experiment setup

A representative cleaning scenario was defined within the living room environment to evaluate the proposed framework. The objective of the simulation was to determine whether the robot could generate an optimal path to visit all target locations, including banana peels and water spills, while avoiding obstacles such as the round table, dining table, chairs, and sofa. A representative simulation experiment was conducted to demonstrate the feasibility of the proposed graph-based approach for generating a collision-free path that enables the robot to complete the cleaning task.

### 5.2 Graph for indoor navigation

Indoor navigation graphs for the cleaning robot were generated, and a representative example is shown in Fig. 4.

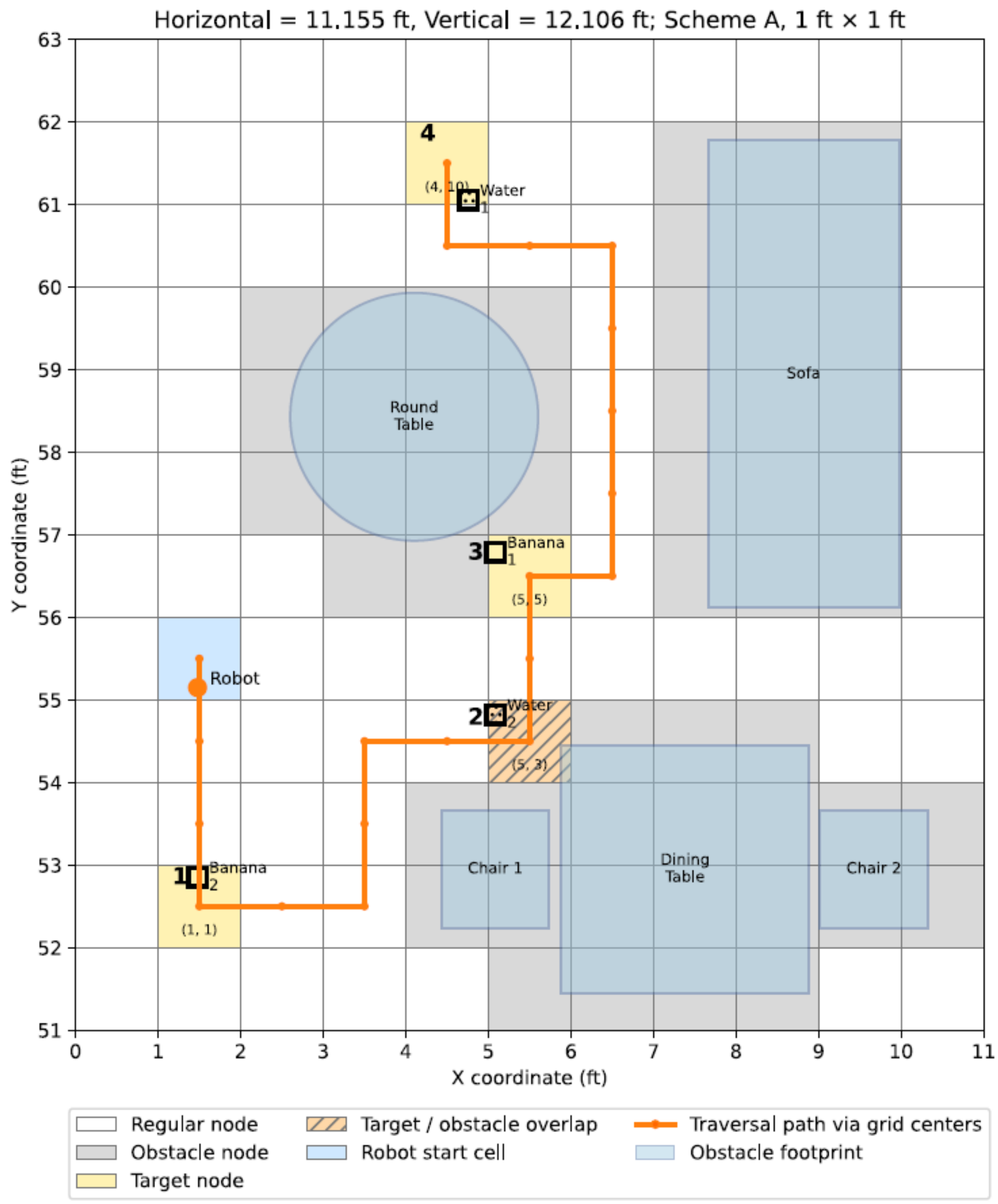

**Fig. 4.** Graph-based indoor navigation for the cleaning robot using a grid resolution of 1 ft × 1 ft

*5.2.1 Graph representation and coordinate system*

To interpret the generated graph, it is necessary to clarify the coordinate system and discretization scheme used in Fig. 4. The coordinate system used in this figure is inherited directly from the original BIM-based spatial data, rather than being redefined as a local coordinate system for the living room. Specifically, the x-coordinates of all elements naturally fall within the range of approximately 0–11 ft, while the y-coordinates fall within 51–63 ft, reflecting their positions in the global building coordinate system. Therefore, the horizontal axis is shown as 0–11, whereas the vertical axis is shown as 51–63, instead of being normalized to a local coordinate system such as 0–12. This choice preserves the consistency between the graph representation and the original BIM data, avoiding additional coordinate transformations. Alternatively, the coordinate system could be shifted to a local frame (e.g., with the origin at the bottom-left corner of the living room), resulting in axes ranging from 0 to the room dimensions. However, such a transformation is not necessary for graph construction, and the original coordinate system is retained for consistency with the source BIM model.

The graph is constructed based on a discretized representation of the indoor environment. The continuous space is partitioned into uniform grid cells of size 1 ft × 1 ft, and each cell is assigned a pair of integer indices (i, j) using a zero-based indexing scheme. A point with continuous coordinates (x, y), such as (5.5, 56.5), is mapped to a grid cell index following the discretization formulation introduced in Section 4.

This formulation generalizes the discretization process and allows flexible adjustment of grid resolution. Therefore, coordinates such as (5, 5) in the graph do not represent exact physical positions, but instead denote grid cell indices in the discretized space. Each node in the graph corresponds to a grid cell, rather than a point in continuous space. This abstraction simplifies graph construction and ensures consistency across obstacle, target, and regular nodes. Furthermore, the discretization facilitates efficient graph-based path planning while preserving a coherent spatial representation of the environment.

### *5.2.2 Graph-based navigation results*

Fig. 4 illustrates a representative graph-based path generated for the cleaning robot under a grid resolution of 1 ft × 1 ft. The graph consists of three types of nodes: regular nodes, obstacle nodes, and target nodes. Obstacle nodes correspond to grid cells intersecting with the projected footprints of furniture elements, while target nodes correspond to the designated navigation destinations of banana peels and water spills.

As shown in Fig. 4, the robot successfully navigates from its initial position and visits all target nodes while avoiding obstacle nodes. The planned path is constructed as a sequence of transitions between adjacent grid cells and is optimized based on traversal cost. The representative result demonstrates the feasibility of the proposed graph representation for supporting efficient and collision-free path planning in indoor environments.

However, due to the coarse grid resolution, certain target-occupied cells overlap with obstacle-occupied cells. This overlap introduces ambiguity in node classification and may lead to infeasible or suboptimal navigation decisions. In this study, obstacle nodes are generally treated as non-traversable; however, when a grid cell is simultaneously classified as a target node, it is considered reachable to ensure that the cleaning task can be completed. This dual classification further highlights the limitation of coarse discretization.

This observation reveals a fundamental limitation of the coarse discretization: the graph representation may become internally inconsistent when a grid cell is simultaneously classified as both obstacle-occupied and target-occupied. In such cases, the modeling assumptions conflict, as obstacle nodes are defined as non-traversable, while target nodes must remain reachable for task completion. This inconsistency not only affects node classification but may also compromise the feasibility and reliability of path planning. Therefore, a refinement of the discretization strategy is required, as discussed in the following section.

### *5.2.3 Resolution refinement and overlap mitigation*

To address the overlap between target-occupied and obstacle-occupied grid cells, the grid resolution was refined by reducing the cell size from 1 ft × 1 ft to 0.5 ft × 0.5 ft. This issue is particularly important because, although obstacle nodes are defined as non-traversable, overlapped cells must still be treated as reachable when they contain targets, which introduces inconsistency in the graph representation.

The overlap issue arises because both obstacles and targets are projected onto the floor plane and mapped to discrete grid cells, which may result in multiple spatial features being assigned to the same cell under coarse discretization. With a finer grid resolution, the spatial representation becomes more precise, allowing obstacle footprints and target locations to be distinguished more accurately. As a result, the overlap between obstacle-occupied and target-occupied cells is significantly reduced or eliminated.

Under the refined discretization, obstacle nodes are defined as grid cells intersecting the projected footprints of obstacles, while target nodes correspond to grid cells containing the designated navigation destinations. For the cleaning scenario considered in this study, these navigation destinations are defined as the centers of banana peels or water spills. The remaining cells are classified as regular nodes. This multi-resolution discretization strategy ensures both geometric fidelity and consistency in graph construction, while maintaining the efficiency of graph-based path planning. This refinement is particularly important for small-scale targets, such as banana peels and water spills, whose spatial extents are significantly smaller than the original grid size.

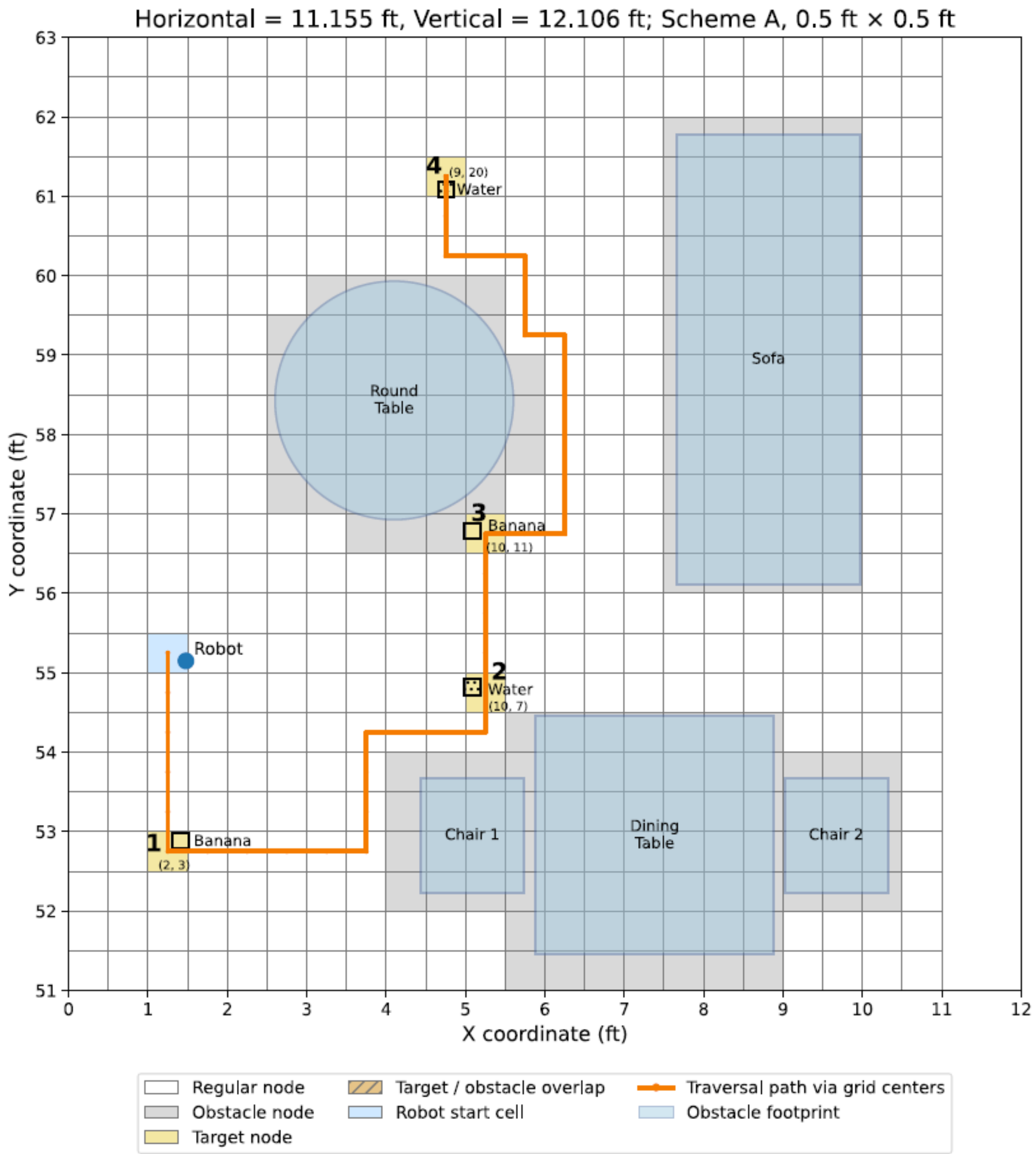

**Fig. 5.** Graph-based indoor navigation for the cleaning robot using a grid resolution of 0.5 ft × 0.5 ft

The updated graph of the cleaning robot's navigation path is shown in Fig. 5. Compared with Fig. 4, the refined grid enhances spatial accuracy and eliminates ambiguity in node classification, resulting in a more reliable graph structure and a more precise navigation path. The presented results demonstrate the feasibility of the proposed graph representation in a representative cleaning scenario. Future work will extend the evaluation to different robot initial positions and more complex indoor environments.

## 6. Result and Discussion

### 6.1 Simulation effectiveness

Fig. 5 presents a representative navigation result generated using the proposed graph-based navigation framework under the refined grid resolution of 0.5 ft × 0.5 ft. Obstacle nodes were defined from the projected footprints of furniture elements (e.g., round tables, dining tables, chairs, and sofas), whereas target nodes corresponded to the designated navigation destinations of banana peels and water spills. For the cleaning scenario considered in this study, these navigation destinations were defined as the centers of the target elements.

As illustrated in Fig. 5, the robot successfully generated a traversal path from its initial position to all target nodes while avoiding obstacle nodes throughout the navigation process. The resulting path demonstrates that the proposed graph representation effectively integrates spatial constraints and task requirements into a unified navigation model. The planned trajectory follows a sequence of grid-cell centers while remaining within regular nodes and target nodes, thereby achieving collision-free navigation. No intersections between the traversal path and obstacle nodes are observed, indicating that the obstacle avoidance strategy is successfully enforced through the graph construction process.

Furthermore, the generated path visits all target nodes through an efficient traversal route under the given discretization. The graph-based search systematically explores feasible paths while avoiding redundant or unsafe movements, demonstrating the feasibility of applying the proposed graph representation to multi-target indoor navigation. Overall, the representative simulation results demonstrate the feasibility of the proposed BIM-enabled graph representation for generating efficient and collision-free navigation paths for indoor cleaning tasks in cluttered indoor environments.

### 6.2 Potential generalization application of the simulation platform

While the simulation experiment presented in this study focuses on a representative cleaning robot scenario, the proposed BIM-enabled, graph-based simulation platform is designed to be generalizable to a wide range of indoor robotic applications. The core framework—comprising spatial discretization, graph construction, and shortest-path optimization—provides a flexible and extensible foundation for modeling and solving various navigation and task allocation problems in built environments.

At its core, the platform transforms a continuous indoor space into a structured graph representation, where nodes correspond to spatial units and edges encode feasible movements with associated traversal costs. This abstraction is independent of the specific task type and can be readily adapted to different operational objectives by modifying node definitions, edge weights, or target configurations. For instance, beyond cleaning tasks, the framework can be applied to inspection scenarios, where robots are required to visit a set of critical building components (e.g., pipes, valves, or sensors) for condition monitoring. Similarly, in facility management, the platform can support maintenance routing, enabling robots or human operators to efficiently traverse multiple service locations while avoiding restricted or hazardous areas. In emergency response scenarios, the graph can be dynamically updated to reflect blocked passages or hazardous zones, allowing the system to recompute safe evacuation or intervention paths in real time. Additionally, the framework can be extended to multi-robot systems, where multiple agents coordinate to complete distributed tasks through graph partitioning or multi-agent path planning algorithms.

The integration of graph search and routing algorithms further enhances the flexibility of the proposed platform. Depending on the application requirements, shortest-path algorithms such as Dijkstra's algorithm or A* can be directly applied to compute collision-free navigation paths, whereas optimization- or heuristic-based routing algorithms (e.g., Traveling Salesman Problem (TSP) formulations) may be incorporated for multi-target navigation. This modular design enables the platform to accommodate different optimization objectives, such as minimizing travel distance, travel time, or energy consumption.

Moreover, the integration of BIM with graph-based navigation introduces new opportunities for incorporating semantic information into robotic decision-making. Unlike traditional grid maps or occupancy grids, BIM models contain rich metadata about building elements, such as functional categories, material properties, and maintenance history. This semantic layer can be leveraged to assign context-aware traversal costs or priorities during path planning. For instance, regions associated with sensitive equipment or restricted access can be assigned higher traversal penalties, while frequently accessed maintenance areas can be prioritized. This capability enables the development of more intelligent and context-aware robotic systems that go beyond purely geometric navigation.

Overall, the proposed simulation platform provides a flexible and extensible framework by decoupling spatial representation from task-specific logic. By leveraging graph theory and optimization algorithms, the framework can be adapted to diverse indoor robotic applications, including cleaning, inspection, maintenance, and emergency management.

### 6.3 Challenges in designing the simulation engine and potential solutions

Designing the proposed simulation engine involved three key challenges: (1) defining the agent and the built environment at an appropriate level of abstraction, (2) integrating the simulation logic with the BIM model, and (3) resolving the overlap between target-occupied and obstacle-occupied grid cells resulting from spatial discretization.

The first challenge lies in defining both the agent and the built environment at an appropriate level of abstraction that supports realistic simulation while remaining computationally tractable. Although the physical characteristics and operational capabilities of robotic agents can be explicitly specified, incorporating all details into the simulation would significantly increase computational complexity and reduce efficiency. Therefore, a balance must be achieved between model fidelity and computational performance. In this study, the cleaning robot is abstracted as a point-based agent navigating through a discretized grid environment, while obstacles are represented through their projected footprints on the floor. This abstraction simplifies the representation of both the agent and the environment while preserving the essential spatial constraints for navigation. As a result, the simulation framework maintains computational efficiency while preserving the essential spatial information required for effective graph-based path planning.

The second challenge involves integrating the simulation engine with the BIM model. This integration was achieved through the use of C# programming and the Revit API, which provide access to both geometric and semantic information within the BIM environment. A key issue in this process is ensuring consistent identification and communication between the simulation code and BIM elements. To address this, a standardized naming convention was adopted, where each simulated entity (e.g., robot, obstacles, and targets) is assigned a unique identifier that is consistently referenced in both the C# code and the BIM model. This enables the simulation engine to correctly interpret and manipulate BIM elements, ensuring synchronized behavior between the digital model and the simulation logic.

The third challenge arises from the discretization of the continuous environment into grid cells. When a coarse grid resolution (e.g., 1 ft × 1 ft) is used, it is possible for target-occupied cells (e.g., containing banana peels or water spills) to overlap with obstacle-occupied cells due to limited spatial resolution. This creates ambiguity in node classification and may lead to infeasible path planning. To resolve this issue, a finer grid resolution (e.g., 0.5 ft × 0.5 ft) is adopted. The refined discretization improves the spatial fidelity of the environment representation, allowing obstacle footprints and target locations to be distinguished more accurately. As a result, the overlap between obstacle and target nodes is significantly reduced or eliminated, leading to more reliable graph construction and navigation outcomes.

Overall, these challenges highlight the trade-offs between modeling accuracy, computational efficiency, and system integration. The proposed solutions demonstrate that careful abstraction, standardized data integration, and appropriate discretization strategies are essential for developing a robust and scalable simulation engine for indoor robotic applications.

**6.4 Limitations of this research and future directions**

Despite the promising results demonstrated in this study, several limitations should be acknowledged.

The identification of target nodes and obstacle nodes currently relies on manual definition for each BIM model. More importantly, the classification of building elements as targets or obstacles is task-dependent and varies across different facility management applications. For example, in cleaning tasks, trash elements such as banana peels and water spills are defined as targets, while furniture elements are treated as obstacles. In contrast, for inspection tasks, the target elements and obstacles may correspond to entirely different categories of building components. This task-specific variability makes manual configuration time-consuming and limits the scalability of the simulation platform across diverse BIM models and facility management scenarios. To address this limitation, future work can introduce a task-oriented element classification library, in which typical facility management tasks are associated with predefined categories (e.g., BIM families) of building elements as targets or obstacles. For a given task, the system can automatically filter relevant building elements from the current BIM model based on their family types or semantic attributes, thereby dynamically identifying target and obstacle nodes without manual intervention. In addition, this approach can be further enhanced by incorporating automated object recognition techniques, such as computer vision or semantic segmentation, to identify targets and obstacles from BIM annotations or other structured data sources. When extended to integrated digital twin systems, these techniques can also leverage real-world sensor data, enabling a more adaptive and scalable simulation framework.

The current approach is based on a grid-based discretization of the environment, which introduces a trade-off between spatial resolution and computational efficiency. While finer grid resolutions (e.g., 0.5 ft × 0.5 ft) improve the accuracy of spatial representation and reduce overlap issues between targets and obstacles, they also increase the size of the graph and computational cost of path planning. Future research may explore adaptive or hierarchical discretization strategies, where grid resolution varies depending on spatial complexity, to balance accuracy and efficiency.

The simulation assumes static environments, where obstacles and targets remain fixed during the planning process. In real-world scenarios, indoor environments are often dynamic, with moving objects or changing conditions that may affect navigation. Extending the framework to support dynamic updates of the graph and real-time path replanning would significantly enhance its applicability to practical robotic systems. The robot is modeled as a simplified point-based agent without explicitly considering its physical dimensions, kinematic constraints, or sensing uncertainties. Although this abstraction facilitates graph construction and path planning, it may limit the realism of the simulation when deployed in real robotic systems. Future work can incorporate more detailed robot models, including footprint constraints and motion dynamics, to improve the fidelity of simulation results. Finally, the current study focuses on a single-agent scenario. In many practical applications, multiple robots may be deployed simultaneously to improve efficiency. Future research can extend the proposed framework to multi-agent systems, incorporating coordination mechanisms and multi-agent path planning algorithms to handle task allocation and collision avoidance among multiple robots.

Overall, addressing these limitations will contribute to the development of a more autonomous, scalable, and realistic BIM-enabled simulation platform for indoor robotic applications.

## 7. Conclusions

This study presents a BIM-enabled, graph-based simulation framework for indoor robotic navigation and task execution. By integrating spatial information from BIM models with graph theory and shortest-path optimization algorithms, the proposed approach enables systematic representation and efficient navigation in complex indoor environments.

The indoor space is discretized into grid cells, which are further classified into obstacle nodes, target nodes, and regular nodes based on the projected footprints of building elements and task-related objects. This unified graph representation allows the robot to perform multi-target navigation while ensuring collision-free movement. The representative simulation results demonstrate the feasibility of the proposed framework for generating efficient and collision-free navigation paths that enable the robot to visit all target locations in the representative cleaning scenario while avoiding obstacles. Furthermore, the study addresses key challenges in simulation engine design, including the integration of C# code with BIM models, the definition of agents and environments, and the resolution of discretization-induced overlap between target and obstacle nodes. By adopting a refined grid resolution, the framework improves spatial fidelity and enhances the reliability of graph construction and path planning.

Beyond the representative cleaning robot case study, the proposed platform provides a flexible and extensible framework that can be adapted to a wide range of indoor applications, including inspection, maintenance routing, and emergency response. The modular structure of the framework allows the integration of various graph-based optimization algorithms, making it adaptable to different task requirements and operational constraints. Despite its effectiveness, the current framework has several limitations, including the need for manual definition of targets and obstacles, the reliance on grid-based discretization, and the assumption of static environments. Future research will focus on automating environment perception, improving dynamic adaptability, and incorporating more realistic robot models and multi-agent coordination strategies.

Overall, this work contributes to the development of intelligent, building-aware robotic systems by providing a structured and extensible simulation platform that bridges BIM-based digital environments and graph-based decision-making.

## 8. Declaration

### 8.1 Author's Contribution

**Xu P:** Coding design and implementation, experiment design, and manuscript writing.
**Gao X:** Manuscript structuring, supervision, and overall guidance of the study.

### 8.2 Conflicts of Interest

The authors declare no conflicts of interest.

### 8.3 Availability of Data and Materials

The BIM model used in this study and the custom-developed Revit plugin code supporting the simulation framework are available from the corresponding author upon reasonable request. The BIM model and the Revit plugin jointly enable the extraction of spatial information from the built environment, including the coordinates of nodes in the graph representation for indoor navigation of the cleaning robot. These data are generated within the simulation platform based on the integration of BIM-based geometric and semantic information.

Due to potential restrictions related to project ownership and software dependencies, the data and code are not publicly available at this time but can be provided for academic and research purposes upon request.

## References

[1] S. Ahn, T. Kim, Y. J. Park, and J. M. Kim, "Improving Effectiveness of Safety Training at Construction Worksite Using 3D BIM Simulation," *Advances in Civil Engineering*, vol. 2020, 2020, doi: 10.1155/2020/2473138.

[2] W. Torres-Calderon, Y. Chi, F. Amer, and M. Golparvar-Fard, "Automated mining of construction schedules for easy and quick assembly of 4D BIM simulations," in *ASCE International Conference on Computing in Civil Engineering 2019*, American Society of Civil Engineers Reston, VA, 2019, pp. 432–438.

[3] C. T. Pérez and D. Bastos Costa, "Increasing production efficiency through the reduction of transportation activities and time using 4D BIM simulations," *Engineering, Construction and Architectural Management*, vol. 28, no. 8, pp. 2222–2247, 2021.

[4] M. A. Alzarrad, G. P. Moynihan, A. Parajuli, and M. Mehra, "4D BIM Simulation Guideline for Construction Visualization and Analysis of Renovation Projects: A Case Study," *Front. Built Environ.*, vol. 7, Mar. 2021, doi: 10.3389/fbuil.2021.617031.

[5] U. Rüppel and K. Schatz, "Designing a BIM-based serious game for fire safety evacuation simulations," *Advanced Engineering Informatics*, vol. 25, no. 4, pp. 600–611, Oct. 2011, doi: 10.1016/j.aei.2011.08.001.

[6] J. B. Kim, W. Jeong, M. J. Clayton, J. S. Haberl, and W. Yan, "Developing a physical BIM library for building thermal energy simulation," *Autom. Constr.*, vol. 50, no. C, pp. 16–28, Feb. 2015, doi: 10.1016/j.autcon.2014.10.011.

[7] S. Kota, J. S. Haberl, M. J. Clayton, and W. Yan, "Building Information Modeling (BIM)-based daylighting simulation and analysis," *Energy Build.*, vol. 81, pp. 391–403, 2014, doi: 10.1016/j.enbuild.2014.06.043.

[8] C. A. Osorio-Sandoval, W. Tizani, E. Pereira, J. Ninić, and C. Koch, "Framework for BIM-Based Simulation of Construction Operations Implemented in a Game Engine," *Buildings*, vol. 12, no. 8, Aug. 2022, doi: 10.3390/buildings12081199.

[9] O. Hosseini and M. Maghrebi, "Risk of fire emergency evacuation in complex construction sites: Integration of 4D-BIM, social force modeling, and fire quantitative risk assessment," *Advanced Engineering Informatics*, vol. 50, Oct. 2021, doi: 10.1016/j.aei.2021.101378.

[10] D. B. West, *Introduction to graph theory*, vol. 2. Prentice hall Upper Saddle River, 2001.

[11] J. A. Bondy and U. S. R. Murty, *Graph theory with applications*, vol. 290. Macmillan London, 1976.

[12] K. Magzhan and H. M. Jani, "A review and evaluations of shortest path algorithms," *Int. J. Sci. Technol. Res*, vol. 2, no. 6, pp. 99–104, 2013.

[13] P.-Y. Gilliéron, D. Büchel, I. Spassov, and B. Merminod, “Indoor navigation performance analysis,” in *Proceedings of the 8th European navigation conference GNSS*, 2004.
[14] C. S. Jensen, H. Lu, and B. Yang, “Graph model based indoor tracking,” in *2009 Tenth International Conference on Mobile Data Management: Systems, Services and Middleware*, IEEE, 2009, pp. 122–131.
[15] W. Liu *et al.*, “Indoor navigation with virtual graph representation: Exploiting peak intensities of unmodulated luminaries,” *IEEE/ACM Transactions on Networking*, vol. 27, no. 1, pp. 187–200, 2018.
[16] M. Fu and R. Liu, “An approach of checking an exit sign system based on navigation graph networks,” *Advanced Engineering Informatics*, vol. 46, p. 101168, 2020, doi: https://doi.org/10.1016/j.aei.2020.101168.
[17] D. Schallmo, C. A. Williams, and K. Lang, “An integrated design thinking approach-literature review, basic principles and roadmap for design thinking,” in *ISPIM Innovation Symposium*, The International Society for Professional Innovation Management (ISPIM), 2018, pp. 1–18.
[18] H. Sayama, *Introduction to the modeling and analysis of complex systems*. Open SUNY Textbooks [Imprint], 2015.
[19] R. Sacks, C. Eastman, G. Lee, and P. Teicholz, *BIM handbook: A guide to building information modeling for owners, designers, engineers, contractors, and facility managers*. John Wiley & Sons, 2018.
[20] Revit, “My First Revit Plug-in Overview,” Autodesk Revit. Accessed: Feb. 12, 2023. [Online]. Available: https://knowledge.autodesk.com/search-result/caas/simplecontent/content/my-first-revit-plug-overview.html#:~:text=Autodesk%20Revit%20has%20a%20.,the%20power%20of%20the%20underlying%20.